\documentclass{article}
\usepackage{arxiv}
\usepackage{pgfplots}
\pgfplotsset{compat=1.17}
\usepackage{booktabs}
\usepackage{caption}

\usepackage{cite}
\usepackage{amsmath,amssymb,amsfonts}
\usepackage{algorithmic}
\usepackage{graphicx}
\usepackage{textcomp}
\usepackage{xcolor}
\usepackage{multirow}
\def\BibTeX{{\rm B\kern-.05em{\sc i\kern-.025em b}\kern-.08em
    T\kern-.1667em\lower.7ex\hbox{E}\kern-.125emX}}

\usepackage{balance}


\begin{document}

\title{Character Detection using YOLO for Writer Identification in multiple Medieval books%\\
%{\footnotesize \textsuperscript{*}Note: Sub-titles are not captured for https://ieeexplore.ieee.org  and should not be used}
}

\author{
  Alessandra Scotto di Freca \\
  Tiziana D'Alessandro \\
  Francesco Fontanella \\
  Filippo Sarria \\
  Claudio De Stefano \\[0.5em]
  \textit{Department of Electrical and Information Engineering (DIEI)} \\
  \textit{University of Cassino and Southern Lazio, Cassino (FR), Italy} \\[0.3em]
  \texttt{\{a.scotto, tiziana.dalessandro, fontanella\}@unicas.it} \\
  \texttt{fi.sarria4901@gmail.com, destefano@unicas.it}
}
\maketitle

\begin{abstract}
Paleography is the study of ancient and historical handwriting, its key objectives include the dating of manuscripts and understanding the evolution of writing. Estimating when a document was written and tracing the development of scripts and writing styles can be aided by identifying the individual scribes who contributed to a medieval manuscript. 
Although digital technologies have made significant progress in this field, the general problem remains unsolved and continues to pose open challenges.
Very interesting results have been obtained in cases of highly standardized book typologies, where the analysis of basic layout features has allowed high recognition accuracy. 
However, these layout-based methods are not very general, as their effectiveness often decreases with texts that follow different styles.
To address the limitations of layout-based methods, we previously proposed an approach focused on identifying specific letters or abbreviations that characterize each writer. 
In that study, we considered the letter ``a", as it was widely present on all pages of text and highly distinctive, according to the suggestions of expert paleographers. We used template matching techniques to detect the occurrences of the character ``a'' on each page and the convolutional neural network (CNN) to attribute each instance to the correct scribe. 
Moving from the interesting results achieved from this previous system and being aware of the limitation of the template matching technique, which requires an appropriate threshold to work, we decided to experiment in the same framework with the use of the YOLO object detection model to identify the scribe who contributed to the writing of different medieval books. 
We considered the fifth version of YOLO, to implement the YOLO object detection model, which completely substituted the template matching and CNN used in the previous work.
The experimental results demonstrate that YOLO effectively extracts a greater number of letters considered, leading to a more accurate second-stage classification. Furthermore, the YOLO confidence score provides a foundation for developing a system that applies a rejection threshold, enabling reliable writer identification even in unseen manuscripts.

%\keywords{Handwriting Recognition, Writer Identification, Object Detection, Convolutional Neural Networks.}
\end{abstract}

\section{Introduction}

Paleography is the scholarly study of ancient and historical handwriting with a primary focus on deciphering, analyzing, and dating manuscripts and documents from past civilizations. This discipline plays a crucial role in understanding the development of writing systems and scripts over time, and it is essential for interpreting historical texts that were written in scripts, languages, or dialects that may no longer be in everyday use.
Paleographers examine a wide range of features within a manuscript, including the shapes and styles of letters, abbreviations, ligatures, punctuation marks, and layout conventions. By analyzing these elements, scholars can identify when and where a document was produced, as well as the social, cultural, or institutional context in which it was written. This work often involves tracing the evolution of scripts across centuries and geographic regions and distinguishing between different scribal hands {\cite{SpecialIssue07,Schomaker2016,paper_2}}.

A very important aspect of paleographic studies is the identification of the different scribes who contributed to the production of an ancient text {\cite{ICPR-2014-Dahllof,Fagioli23,Gatta2023}}.
In recent years, this field has seen a growing integration of digital technologies, significantly enhancing traditional palaeographic methods {\cite{stokes09,Stokes15,paper_4, CILIA202339}}.
Techniques such as image processing, machine learning, and handwriting pattern and style recognition have been increasingly adopted, allowing scholars to analyze manuscripts at a scale and depth previously unattainable with manual approaches{\cite{paper_5,paper_6}}.
However, the problem of identifying the different hands that produced an ancient text is still far from being solved in the general case and still represents one of the most difficult challenges to face.
As discussed in \cite{CILIA2020}, techniques for identifying scribes in ancient manuscripts generally fall into two main categories.
The first includes methods that focus on analyzing individual letters, symbols, or abbreviations, either from single lines of text or entire pages. \cite{paper_1,paper_3}. These approaches rely heavily on the ability to accurately segment the manuscript text into individual letters or graphemes: an inherently challenging task that often yields unreliable or suboptimal results \cite{Bulacu2007}.
The second category includes techniques that extract features from the entire manuscript page. These techniques use texture or layout features and have shown particularly promising results in cases involving highly standardized handwriting and book typologies. In such contexts, analyzing layout features regarding the page organization and how the scribe utilized the available space, may provide valuable insights for distinguishing between very similar handwriting styles, even without relying on traditional paleographic analysis \cite{Marinai2020}.
In previous studies \cite{DeSte18EAAI,CILIA2020_1,DeSte18MEAS}, we proposed several pattern recognition systems aimed at distinguishing between the scribes who collaborated on the transcription of a single medieval Latin manuscript. 
These systems employed a carefully selected set of features derived from page layout analysis, informed by insights from palaeographic research, and utilized standard machine learning techniques for classification. 
We also explored deep learning approaches, proposing a deep transfer learning method for row detection and page classification, which yielded highly promising results \cite{CILIA2020, JIMG2020}.

A key limitation of layout-based approaches is their poor transferability across manuscripts, making it difficult to identify the same scribe in texts created with different standards.
To overcome this limitation, in \cite{MT_icpr_25} we proposed a preliminary approach that relies on paleographic expertise, focusing on letters or abbreviations generally recognised as distinctive to identify individual scribes. 
We selected the letter ``a" as the reference symbol, widely present in manuscript pages and recognized by paleographers as one of the most distinctive symbols. Our analysis, conducted on two medieval books, namely the Avila Bible and the Trento Bible, involved extracting occurrences of the letter ``a" from binarized images of the pages.
This process, denoted as Template Matching (TM), involved sliding a template over a page image to locate matching regions based on a similarity score\cite{CCN_tempMatc}. A threshold was applied to select only reliable matches, from which letter occurrences were extracted and saved. These were then cleaned using thinning and contour detection to remove traces of nearby characters and given as input to the CNNs.
To ensure accurate extraction of the target letters and facilitate effective postprocessing, we chose to exclude noisy or unclear images. This decision proved significant, as the high accuracy achieved when analyzing isolated, clean samples was not replicated when the same method was applied to entire manuscript pages.
These samples were then processed using convolutional neural networks (CNNs) to train a classifier capable of assigning each ``a" to the correct scribe. Finally, we applied a majority voting strategy to attribute each manuscript page to the scribe with the highest number of identified occurrences of the character.
However, despite the excellent performance achieved by training CNNs at the sample level, TM often fails to extract samples of the letter ``a", thus reducing the accuracy at the page level. This behavior is mainly due to the difficulty of finding effective threshold values in the TM algorithm.

% the effectiveness of this approach decreased when applied to entire pages. Medieval manuscripts, in fact, often suffer from image quality and text integrity issues due to material degradation (e.g., stains, holes), inconsistent digitization conditions, ink bleed-through, and handwriting variability. These challenges complicate text analysis and character extraction, limiting the effectiveness of automated recognition methods. 
% Io eliminerei questo pezzo ---------------------------------------------------------------
%In the previous approach, the process involved sliding a template over a page image to locate matching regions based on a similarity score\cite{CCN_tempMatc}. A threshold was applied to select only reliable matches, from which letter occurrences were extracted and saved. These were then cleaned using thinning and contour detection to remove traces of nearby characters and given as input to the CNNs.
%To ensure accurate extraction of the target letters and facilitate effective postprocessing, we chose to exclude noisy or unclear images. This decision proved significant, as the high accuracy achieved when analyzing isolated, clean samples was not replicated when the same method was applied to entire manuscript pages. 
%-------------------------------------------------------------------------------
Moving from the above considerations, we tried to overcome the limitations previously discussed by training a neural network directly on the manuscript page. The goal is to enable the model to identify and learn the distinctive features of each scribe by focusing on the characteristic letters present throughout the page.
To address this need, we choose to use YOLO as an object detection tool in its version YOLOv5s6\cite{JIANG20221066}.
Recent studies on layout analysis of historical document images have demonstrated YOLO's latest versions to effectively detect and localize elements within complex, degraded, or variably formatted manuscripts \cite{YoloCH}.

%forse questa parte andrebbe spostata
%In \cite{YOLO_CH_TL} YOLO has been adapted to detect text blocks, illustrations, and other structural components, facilitating tasks such as automated transcription, layout understanding, and preservation efforts. Transfer learning techniques have also been applied to tailor YOLO models trained on modern datasets to the specific visual characteristics of ancient documents, improving detection accuracy without extensive manual annotation.
%b Eliminerei questa frase: Considering the YOLO’s significant role in advancing the automated processing of ancient documents, we decide to use it for special character extraction from the page images. 

In this preliminary study, the first aim is to understand whether YOLO could extract more characters with respect to the method previously used. Therefore, our analysis started from the extraction of the same letter ``a'' used in \cite{MT_icpr_25}. The manuscript pages were segmented into columns, binarized, and then YOLO was used to extract the characters.

% ELIMINATA ---------------------------------
%Additionally, we aim to exploit YOLO as a classifier, analyzing the class probability of each sample to attribute a page to a specific writer. In particular, we performed two experiments. In the first experiment, we performed an iterative training of YOLO to recognize the higher number of sample instances in one document.  The iterative process ends when all pages of the first document are used, and the highest number of samples is found. In the second experiment, we used the samples identified in the first experiment to train YOLO to recognize the samples written only by the author, who we know contributed to both documents. Thus we train YOLO on one document and use the model to infer the presence of the author on the other document.
%-------------------------------------------------------------------------------------------------
The results obtained were very interesting, showing performance higher than those obtained previously.
The main contributions are summarized in the following 
\begin{itemize}
    \item providing a numerically richer annotation of instances compared to the previous work, significantly increasing the amount of data available for object detection of specific characters;
    \item demonstrating that fine-tuning a YOLO-based object detector on instances of a character written by a specific scribe results in a model capable of detecting the same characters as in another manuscript. The confidence scores (CSs) of these detections reflect varying degrees of similarity to the target scribe’s writing style;
    \item discussing the potential of defining a scribe-specific attribution system by determining a recognition threshold based on fine-tuning and evaluation within a single manuscript. This lays the basis for future work on personalised writer identification;
    %\item showing that it is possible to adjust the detection threshold to influence the trade-off between accuracy and error. The results presented in our threshold analysis support this flexible approach and highlight its potential for controlling the level of similiarity in character detection.
\end{itemize}
The remainder of the paper is organized as follows: Section 2 illustrates the datasets derived from both the Avila and the Trento bibles and YOLO description, Section 3 describes the proposed method discussing the different parts in which it is articulated, while the experimental results are presented in detail in Section 4. Discussion and future works are eventually left to Section 5.

\section{Materials}
\label{sec:materials}

\subsection{Manuscripts and Data Description}
\label{sec:dataset}
In this study, we used two Medieval Bibles, namely the Avila Bible and the Trento Bible.
In medieval books, pages are typically written on both sides of a single sheet. When referring to individual pages in these documents, the terms "recto" and "verso" are used instead of front and back, respectively. 

The Avila Bible was written in Italy in the early 12th century by at least nine different scribes \cite{7}. It was later taken to Spain, where local scribes completed the text and added decorative elements. A third stage of work occurred in the 15th century when another scribe contributed additional material.
%Because it features contributions from multiple scribes, both contemporary and from later periods, the Avila Bible offers an excellent benchmark for evaluating automatic scribe identification systems. To the best of our knowledge, no other publicly available dataset combines such high-quality reproductions with a manageable number of recurring, identifiable hands.
The Avila Bible consists of 870 two-column pages, though for the purposes of this study, pages of particularly poor quality were excluded. In particular, from the initial section of the Bible, we considered 432 images from the recto side and 430 from the verso side.
Paleographic analysis has identified at least 13 distinct scribal hands within the manuscript. One of these appears to have contributed only decorative letters, which were removed during preprocessing; as a result, this scribe was excluded from the identification process.
In addition, paleographers provided guidance on the distinctive letterforms used specifically by Italian scribes active in the 12th century. Based on this, we excluded pages written by 15th-century scribes and Spanish contributors, focusing solely on the contemporary Italian hands.
Scribes varied in their level of contribution: while some wrote only a single page, others authored substantial portions—up to 143 pages. This uneven distribution of samples creates a significantly imbalanced classification problem.
High-resolution images of the manuscript are freely accessible online \cite{AvilaB}.
Pages were digitized at a resolution of $4100\times6110$ pixels and were individually labeled by an expert paleographer, using the letters A, B, C, D, E, F, G, H, and I to represent the corresponding writers.

%To support the data requirements of deep learning algorithms, we excluded scribes with too few pages. Our analysis focused on a curated subset of 705 pages, each containing clearly identifiable handwriting from one of eight selected scribes. These pages were digitized at a resolution of 4100×6110 pixels and were individually labeled by an expert paleographer, using the letters A, B, D, E, F, G, H, and I to represent the corresponding writers.

The Trento Bible is an Atlantic volume dating back to the first half of the 12th century.
Unlike the more complex history of the Avila Bible, the Trento Bible was produced within a narrower time frame and has not undergone later additions or alterations. 
The surviving portion of the manuscript has been thoroughly examined by palaeographic experts. The decorative elements are largely attributed to a single individual, known as the Master of the Avila Bible, along with his workshop. Furthermore, palaeographers have identified three distinct scribes responsible for the transcription of the Trento Bible. Notably, one of these scribes (referred to as Scribe F) also contributed to the Avila Bible, establishing a clear link between the two manuscripts.
Although the entire Trento Bible has been digitized, the number of high-quality images available for analysis remains limited. The manuscript consists of 394 pages, each digitized at a resolution of  $2832\times4256$  pixels.
An additional main difference with Avila is that Trento's bible pages follow two different formats: two-column and three-column.
In particular, 67 pages (34 recto and 33 verso) have the three-column format, and 327 (163 recto and 164 verso) belong to the two-column format.
% Figure \ref{Fig:AvilaTrento} illustrates a visual comparison between pages from the Avila and Trento Bibles.

\subsection{YOLO}
\label{sec:yolo}
YOLO is real-time object detection, that models object detection as a single regression problem, directly predicting bounding boxes and class probabilities from full images using a single neural network. Its first version was proposed for detecting objects in video frames, and it had some limitations in localizing small objects\cite{YOLO_1}. In its subsequent versions, YOLO improves on various aspects such as speed, accuracy, generalization, support for small object detection, and scalability. %The YOLO version used in our system is the fifth, written in PyTorch and widely used due to ease of use and training. Among YOLO versions, the v5 uses the EfficientDet architecture that is based on the EfficientNet network architecture. Additionally, uses a new method for generating the anchor boxes, called "dynamic anchor boxes." It involves using a clustering algorithm to group the ground truth bounding boxes into clusters and then using the centroids of the clusters as the anchor boxes. This allows the anchor boxes to be more closely aligned with the detected objects' size and shape.
%Finally, YOLOv5 introduces the concept of "Spatial Pyramid Pooling" (SPP), a type of pooling layer used to reduce the spatial resolution of the feature maps. SPP is used to improve the detection performance on small objects, as it allows the model to see the objects at multiple scales. This last aspect of the v5 was particularly interesting for our system, where the samples of special characters are very small with respect to the entire handwriting space. For a deeper understanding of the used model, YOLOv5s6, it is possible to read 
Our system uses YOLOv5, developed in PyTorch, which is popular for its ease of use and training. YOLOv5 integrates a backbone that extracts feature maps at multiple scales, which are then merged by the feature fusion network (neck). This process produces three feature maps at resolutions of
80×80, 40×40, and 20×20 used to detect small, medium, and large objects, respectively. These maps are passed to the prediction head, where each pixel is evaluated using preset anchor boxes to compute object confidence and perform bounding box regression, resulting in a multi-dimensional array containing class labels, confidence scores, and bounding box parameters\cite{Liu2022-wh}.
%integrates components like the EfficientDet architecture (built on EfficientNet) and a novel dynamic anchor box generation method that uses clustering to better match object shapes and sizes.
It also uses Spatial Pyramid Pooling\cite{SPP_15}, a technique that enhances detection performance on small objects by enabling the model to analyze features at multiple scales.
This last feature is especially beneficial in our context, where the special characters to be identified are small compared to the full handwritten page.
For training the models, the YOLOv5s6\cite{YOLOv5s6} architecture was used, a lightweight variant of the YOLOv5 family, particularly well-suited for high-resolution images. The downloaded YOLOv5s6 architecture was pre-trained on COCO dataset\cite{coco}, and used our system with different fine-tuning strategies described in section \ref{sec:method}. The fine-tuning of YOLOv5s6 consists of adapting the pre-trained model to our dataset using the page images and corresponding annotations providing the class identifier and bounding box of characters.
The more character annotations there are, the better the trained model. 
After training, the model can be used for inference to predict the bounding boxes of similar characters in new images. For each detected bounding box, the model also provides the predicted class label and a confidence score (CS). In our study, class label 1 is assigned to instances of the letter  ``a"  written by the scribe who contributed to both manuscripts, while label 0 is used for all other occurrences of the letter  ``a".

\section{The Proposed Method}
\label{sec:method}
\begin{figure*}[t]
    \centering
    \includegraphics[width=0.9\textwidth, trim={0cm 0.5cm 0cm 4cm}, clip]{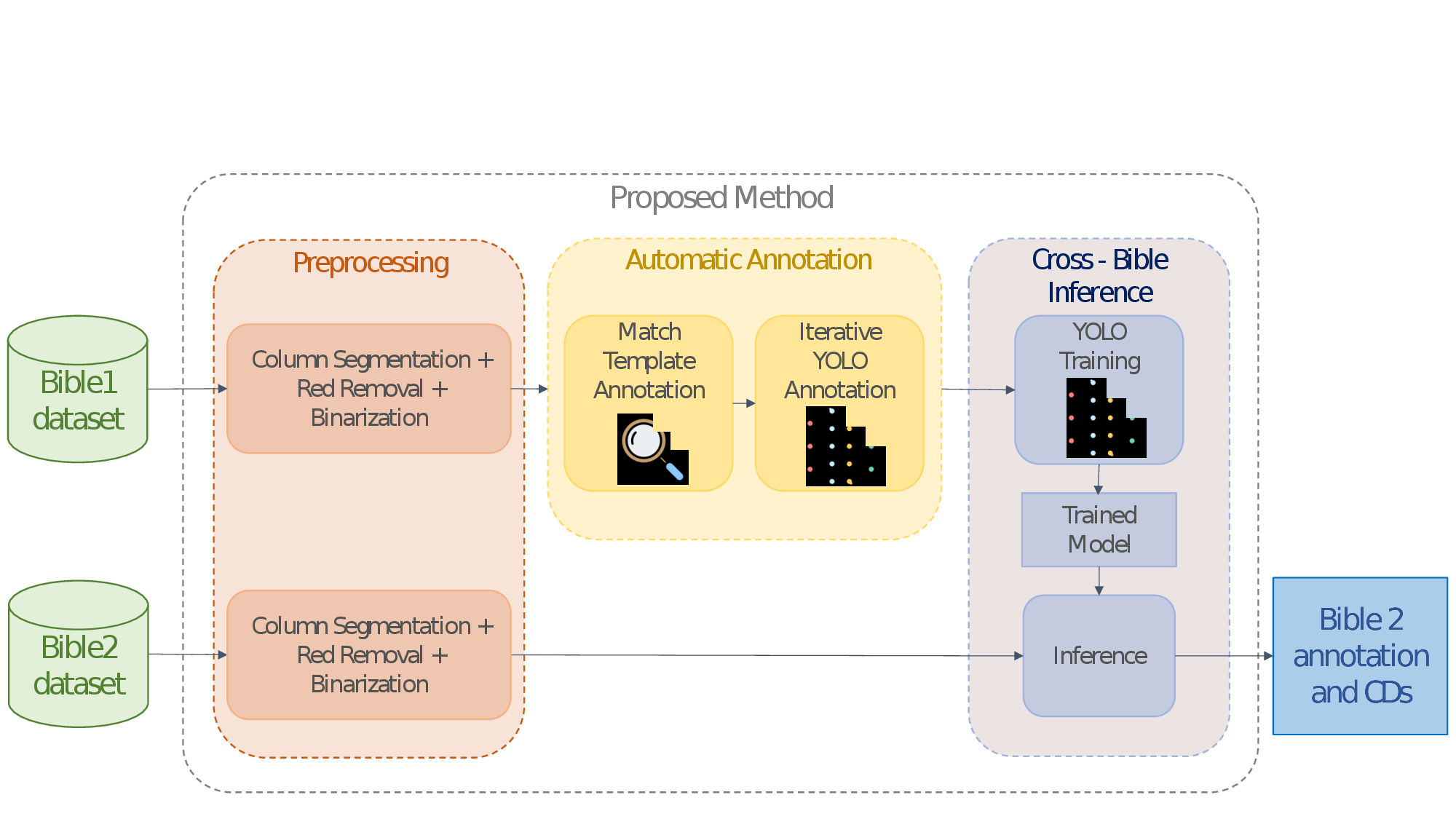}
	\caption{Proposed method.}
	\label{fig:method}
\end{figure*}

This work presents a structured pipeline to train a YOLO-based object detection model for identifying the letter “a” in digitised images of medieval manuscripts. The aim is to detect instances of the letter produced by a specific scribe (Scribe B in the Trento Bible, also known as Scribe F in the Ávila Bible) and apply the learned model to unseen documents for character detection, stylistic analysis or scribal attribution. The proposed method is depicted in  Figure \ref{fig:method} and divided into key stages described in the next sections.

\subsection{Preprocessing}
\label{sec:preprocessing}
The bible pages analysed in this study were digitised at high resolution to ensure both textual clarity and minimal visual degradation. Despite the overall image quality, preprocessing was necessary to support the implementation of segmentation and deep learning (DL) techniques. 
\begin{itemize}
    \item Column segmentation: The segmentation of text columns was performed by manually determining the Regions of Interest (ROIs) corresponding to the columns present in the pages. Specifically, the coordinates and dimensions of these ROIs were identified on a representative sample of pages and differ from one manuscript to another. Once defined, these cropping parameters were uniformly applied to all pages, ensuring consistent extraction of column content across the entire collection.
    \item Red removal: The original images were in RGB format and typically featured three main colour components: the background (often textured), the main body of text, and decorative initials or symbols frequently rendered in red. To ease the following binarisation process, a preliminary step was introduced in which all red pixels were converted to white. This adjustment improved text legibility and simplified the subsequent processing stages.
    \item Binarisation: Following the removal of red elements, images were first converted to grayscale to normalise intensity levels, then binarisation was performed using Otsu’s thresholding algorithm\cite{Otzu}.
\end{itemize}
Through these preprocessing operations, the digitised images were transformed into a standard format suitable for advanced analysis, facilitating both the identification of textual elements and the application of DL-based recognition methods. The described procedure was applied to both the ancient bibles.

\subsection{Automatic Annotation}
\label{sec:annotation}
The output of the preprocessing phase consists of binarised column images of the medieval manuscripts considered. One of the objectives of this study is to automatically annotate all instances of the letter “a” within these column image. 
In this context, “annotation” refers to the process of automatically identifying and localising each occurrence of the character “a” by drawing a bounding box around it. These bounding boxes define the spatial coordinates (top-left x, y, width, height) of the detected character within the image.
This is achieved on the Trento Bible using a structured, iterative training based on the YOLO object detection model. YOLO processes the entire image and predicts the coordinates of the bounding box for each detected object, the class label (in our case, whether the detected shape is the letter “a”) and a CS indicating the model’s certainty about the detection.
The iterative annotation process is carried out separately for the Trento bible and consists of the following steps:

\begin{itemize}
\item Initial Template Matching Annotation: A reference image of the target character written by the scribe of interest (B), is provided to a Template Matching (TM) algorithm. This algorithm is used to automatically annotate occurrences of the letter “a” in an initial subset of pages. To achieve this, we used a method provided by the OpenCV library: {the normalised cross-correlation TM function \cite{CCN_tempMatc}}. Manual review is performed to correct any mislabelled instances and ensure high-quality training data.
\item Iterative YOLO-based Annotation: A YOLO model, pretrained on the COCO dataset, is fine-tuned using the first 60 ritto pages (120 columns) written by Scribe B from the Trento Bible, where instances of the letter “a” were initially annotated through TM and manually reviewed for quality. The fine-tuned model is then applied to a separate set of 150 unseen ritto pages to automatically detect and annotate additional occurrences of the letter “a.” These predictions are manually reviewed, and the correctly identified instances are added to the training set. In the next iteration, the model is retrained on this expanded training set (now covering 210 ritto pages) and evaluated on the remaining ritto and verso columns from the Trento manuscript to obtain annotations on the remaining pages of the Trento bible. 
This iterative process—comprising inference, review, and retraining—was conducted over three main training cycles, progressively improving the model’s ability to generalise across pages and subtle handwriting variations. The process concludes once the model demonstrates satisfactory performance and can reliably annotate all columns of the Trento manuscript.
\end{itemize}

At the end of this phase, every column of the Trento bible written by Scribe B has been annotated with instances of the letter “a”, allowing for scalable quantitative analysis across documents and scribes.

\subsection{Cross-Bible Inference}
\label{sec:inference}
With all Trento columns from Scribe B fully annotated, this phase explores the use of the YOLO-based model for generating character annotations on Avila manuscript and evaluate if it is possible to define a scribe attribution rule through cross-manuscript inference. A YOLO model, initially pretrained on the COCO dataset, is fine-tuned exclusively on annotated columns written by the target scribe in the Trento manuscript. This process results in a detector that specialises in detecting occurrences of the letter “a” and matching the stylistic characteristics typical of that scribe’s handwriting.
Unlike a traditional classification task, our approach does not train an explicit writer classifier. Instead, we evaluate the CS associated with each detection, which reflects how closely a detected shape matches the style of the scribe seen during training. Importantly, this value does not merely confirm the presence of the letter “a”, but rather indicates how closely the detected character matches the specific stylistic traits of the target scribe.
During inference, the trained model is applied to the Avila manuscript. The outcome of this step is twofold: first, we obtain automatic annotations of all instances of the letter “a” in Avila; second, we analyse the distribution of CS values to determine whether high-confidence detections correspond to handwriting stylistically consistent with the Trento scribe. By systematically varying the confidence threshold, we demonstrate that it is possible to filter detections and retain only those that exhibit strong stylistic similarity—thus offering a promising mechanism to support scribe attribution across manuscripts.

\section{Experimental Approach and Results}
\label{sec:experimental}
The objective of this section is twofold: first, to provide all the necessary details for understanding the practical application of the proposed method described in Section \ref{sec:method}; and second, to report and analyse the results obtained through its implementation.
Figure \ref{fig:method} details the approach step by step. First of all, both the bibles pages went through preprocessing, as detailed in Section \ref{sec:preprocessing}. Then Trento column images went through the annotation phase to detect the letter ``a" instances. Finally, the YOLO model pretrained on the COCO dataset was fine-tuned on the annotated columns of Trento written by Scribe B. After the training phase, the resulting model was applied to the Avila bible to detect instances of the letter ``a" and check the CSs obtained. 

\subsection{Annotation Results}
\label{sec:ann_results}
To assess the effectiveness of the trained YOLO object detection model and provide a comparative analysis, we extracted occurrences of the letter “a” from the Ávila Bible, using the model trained on the Trento Bible. Table~\ref{table:avila_icpr25extraction} presents the number of extracted instances across multiple scribes (A–I), along with a comparison to the TM method proposed in~\cite{MT_icpr_25}.
The last column provides the percentage increase (or decrease) in the number of instances detected by our method with respect to TM.
Overall, our YOLO method consistently detects a higher number of instances than TM, with improvements ranging from +5\% to over +35\% depending on the scribe. This suggests a superior capability of the YOLO model in identifying graphical units within the pages, potentially due to better generalisation and localisation performance.
However, some variation is observed among different scribes. For example, Scribe C and Scribe F show the largest improvements, indicating that our method is particularly effective for these writers, possibly due to the more consistent or distinguishable visual patterns in their writing. On the other hand, the lowest improvement is observed for Scribe H, suggesting that in this case, the YOLO-based method still performs well but offers a smaller advantage over TM. No cases were found where our method detected fewer instances than TM.

These findings indicate that while our approach generally outperforms the baseline across all scribes, the extent of the improvement can vary depending on individual handwriting characteristics.

\begin{table*}[!t]
\caption{Details about the image extraction process of the letter ``a" from Avila Bible using Trento bible for training YOLO, and comparison with the TM method from \cite{MT_icpr_25}.}
\begin{center}
\begin{tabular}{llcccccccccc}
\hline
\textbf{} && \textbf{A} & \textbf{B} & \textbf{C} & \textbf{D} & \textbf{E} & \textbf{F} & \textbf{G} & \textbf{H} & \textbf{I} & \textbf{Total\_Av} \\
\hline
\multirow{3}{*}{TM} &\#occ. & 2933 & 166 & - & 378 & 565 & 1317 & 269 & 425 & 52 & 6105\\
&\#total\_pages & 309 & 31 & -& 18 & 76 & 150 & 31 & 37 & 53 & 705 \\
&\#occ./page  & 12.02 & 6.91 & - & 23.62 & 7.43 & 8.78 & 38.42 & 12.14 & 1.85 & 13.89 \\\hline
\multirow{4}{*}{YOLO} &\#occ. & 88035 & 8461 & 2330 & 6734 & 23983 & 42756 & 8461 & 6010 & 15524 & 202294
\\
&\#total\_columns  & 3713 & 64 & 23 & 52 & 158 & 318 & 64 & 59 & 110 & 1561 \\
&\#occ./column  & 123.47 & 132.20 & 101.30 & 129.50 & 151.79 & 134.45 & 132.20 & 101.86 & 141.13 & 127.54\\
%&occ\_std\_dev\ & 27.56  & 15.15  & 47.50  & 27.53  & 20.83  & 34.19  & 15.15  & 42.58  & 32.59 &29.23  \\
&conf\_Av          & 0.7854 & 0.7773 & 0.7279 & 0.7169 & 0.8152 & 0.8336 & 0.7773 & 0.6980 & 0.7175 &0.76\\
%&conf\_std\_dev       & 0.1450 & 0.1440 & 0.1723 & 0.1677 & 0.1090 & 0.1016 & 0.1440 & 0.1838 & 0.1811 \\
\hline
\end{tabular}
\end{center}
%\end{adjustbox}
\label{table:avila_icpr25extraction}
\end{table*}

\subsection{Confidence Discussion}
\label{sec:discussion}
This section discusses the outcomes of our confidence-based inference approach for writer attribution. Specifically, we trained the YOLO model on all instances of the letter “a” from Trento pages written by Scribe B, and tested it on the Avila dataset to evaluate the model’s ability to detect “a” shapes that are stylistically similar to those of Scribe B.

To explore this, we varied the confidence threshold produced by the YOLO model on the test set (Avila) from 0.70 to 0.85. Each detected letter “a” is associated with a CS, which measures how closely the shape matches the style of Scribe B, as learned during training. Notably, this is not a classification task in the traditional sense. We did not train a classifier that labels each instance as “B” or “not B.” Instead, we used the CS as a measure of stylistic similarity, and by applying different thresholds, we filtered detections to retain only those most representative of the scribe’s style.

\begin{figure}[!t]
\centering
\begin{tikzpicture}
\begin{axis}[
    width=8.9cm,
    height=6cm,
    xlabel=Threshold,
    xmin=0.7, xmax=0.85,
    ymin=0, ymax=100,
    xtick={0.7,0.72,0.74,0.76,0.78,0.8,0.82,0.84},
    tick label style={font=\scriptsize},
    ytick={0,20,40,60,80,100},
    legend pos=south east,
    grid=both,
    legend style={font=\scriptsize},
    thick,
    legend style={
        font=\scriptsize,
        at={(0.5,-0.25)},
        anchor=north,
        draw=none,
        legend columns=2,
        column sep=10pt
    }
]
\addplot[
    color=blue,
    mark=*,
]
coordinates {
    (0.70,34.79)
    (0.71,35.16)
    (0.72,35.85)
    (0.73,36.58)
    (0.74,37.47)
    (0.75,38.68)
    (0.76,40.27)
    (0.77,42.68)
    (0.78,46.91)
    (0.79,53.41)
    (0.80,63.92)
    (0.81,72.93)
    (0.82,79.76)
    (0.83,81.78)
    (0.84,75.23)
    (0.85,40.80)
};
\addlegendentry{F-Score (\%)}

\addplot[
    color=orange,
    mark=*,
]
coordinates {
    (0.70,26.28)
    (0.71,27.49)
    (0.72,29.63)
    (0.73,31.83)
    (0.74,34.38)
    (0.75,37.65)
    (0.76,41.67)
    (0.77,47.35)
    (0.78,56.25)
    (0.79,66.28)
    (0.80,78.26)
    (0.81,85.75)
    (0.82,90.70)
    (0.83,92.64)
    (0.84,91.23)
    (0.85,84.28)
};
\addlegendentry{Accuracy (\%)}
\end{axis}
\end{tikzpicture}
\caption{Accuracy and F-score values for different confidence thresholds.}
\label{fig:confidence_plot}
\end{figure}
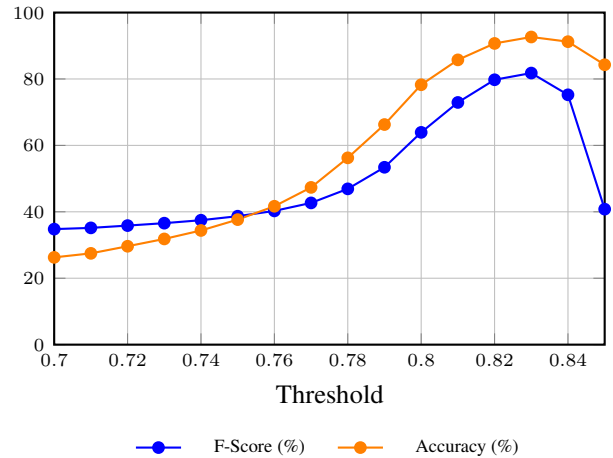

For each threshold setting, we computed two metrics: accuracy (i.e., the proportion of correct attributions and rejections) and error rate (i.e., the combined false positives and false negatives). Figure~\ref{fig:confidence_plot} reports how the accuracy increases with the threshold, reaching a maximum of 92.64\% for a threshold of 0.83, with a corresponding F-Score of 81.78\%. Beyond this threshold, performance deteriorates due to overly aggressive filtering, which discards true positives.

These results show that tuning the threshold enables a flexible trade-off between inclusiveness and precision, allowing us to explore the strength of stylistic matches across manuscripts. While we did not build a classifier, this experiment shows that the confidence threshold itself can serve as a foundation for classification. In future work, we mean to define a writer-specific threshold using data from a single manuscript (e.g., Trento), training the model on all instances, fine-tuning it on Scribe F, and then testing on unseen “F” and “non-F” pages to empirically determine an optimal classification threshold. This threshold could then be applied to infer writer presence in new documents.

Moreover, reasoning on the CSs allows palaeographers to identify authors who share stylistic characteristics, potentially indicating common training backgrounds or scriptoria. High-confidence matches across different texts may suggest that certain scribes were influenced by similar writing schools or traditions, providing valuable insights for historical and palaeographic analysis.

Thus, our results highlight a key contribution: CSs can be used not only for detection but also as decision-making signals for attribution and stylistic grouping, opening the door to threshold-based scribe classification models and comparative palaeographic studies.

\section{Conclusions and future work}
\label{sec:concs}
This study introduced a structured, YOLO-based pipeline for the detection of specific characters in digitised medieval manuscripts, aiming to support both automatic annotation and writer attribution. By focusing on the letter “a”, a graphically distinctive and frequently occurring character, we proposed a multi-phase method to annotate large-scale manuscript data and reason about scribal attribution using CSs.

Our approach achieved several key contributions:
\begin{itemize}
    \item we demonstrated that our YOLO-based system can extract a significantly higher number of character instances compared to the TM approach, providing a richer and more comprehensive annotated dataset for paleographic analysis;
    \item by fine-tuning the model on samples written by a specific scribe, we showed that it is possible to detect stylistically similar characters in a second manuscript, revealing visual patterns consistent with the scribe's handwriting;
    \item we investigated the interpretation of YOLO's CS as a measure for stylistic similarity. This opens up the possibility of using threshold-based reasoning to determine the presence of a specific scribe’s hand across different documents;
    \item we highlighted that, although no explicit writer classifier was trained, manipulating the CS threshold allowed us to balance accuracy and error. Our experiments support the viability of building future writer identification systems based on this thresholding approach.
\end{itemize}

Our experimental results confirmed the effectiveness of this methodology. Not only did the YOLO model consistently outperform the prior TM method in detecting occurrences of the letter “a”, but it also allowed for inference about scribal attribution based on confidence values. Furthermore, the analysis across multiple scribes revealed that certain writers were easier to detect, likely due to the consistency and recognizability of their handwriting styles.

For future work, we intend to formalise the concept of confidence thresholding into a classification system. This involves training the YOLO model on one manuscript, determining an optimal threshold for the CS using in-manuscript validation, and applying it to other documents to infer authorship. Additionally, analysing the CS distribution across multiple scribes could support the identification of stylistic clusters, shedding light on shared training backgrounds or scriptorial traditions. This line of research not only advances the automation of paleographic analysis but also offers new tools for historians to investigate the provenance and authorship of ancient texts.


\begin{thebibliography}{10}

\bibitem{SpecialIssue07}
A.~Antonacopoulos and A.~C. Downton, ``Special issue on the analysis of
  historical documents,'' {\em IJDAR}, vol.~9, no.~2-4, pp.~75--77, 2007.

\bibitem{Schomaker2016}
S.~He, P.~Samara, J.~Burgers, and L.~Schomaker, ``Image-based historical
  manuscript dating using contour and stroke fragments,'' {\em Pattern
  Recognition}, vol.~58, pp.~159 -- 171, 2016.

\bibitem{paper_2}
E.~O. Omayio, S.~Indu, and J.~Panda, ``Historical manuscript dating:
  traditional and current trends,'' {\em Multimedia Tools Appl.}, vol.~81,
  p.~31573–31602, sep 2022.

\bibitem{ICPR-2014-Dahllof}
M.~Dahllof, ``{Scribe Attribution for Early Medieval Handwriting by Means of
  Letter Extraction and Classification and a Voting Procedure for Larger
  Pieces},'' in {\em {Proceedings of the 22nd International Conference on
  Pattern Recognition}}, pp.~1910--1915, IEEE Computer Society, 2014.

\bibitem{Fagioli23}
A.~Fagioli, D.~Avola, L.~Cinque, E.~Colombi, and G.~L. Foresti, ``Writer
  identification in historical handwritten documents: A latin dataset and a
  benchmark,'' in {\em Image Analysis and Processing - ICIAP 2023 Workshops}
  (G.~L. Foresti, A.~Fusiello, and E.~Hancock, eds.), (Cham), pp.~465--476,
  Springer Nature Switzerland, 2024.

\bibitem{Gatta2023}
A.~Gattal, C.~Djeddi, F.~Abbas, I.~Siddiqi, and B.~Bouderah, ``A new method for
  writer identification based on historical documents,'' {\em Journal of
  Intelligent Systems}, vol.~32, no.~1, p.~20220244, 2023.

\bibitem{stokes09}
P.~Stokes, {\em {Computer-Aided Palaeography, Present and Future}},
  pp.~309--338.
\newblock Institut f\"{u}r Dokumentologie und Editorik, 2009.

\bibitem{Stokes15}
P.~A. Stokes, ``Digital approaches to paleography and book history: Some
  challenges, present and future,'' {\em Frontiers in Digital Humanities},
  vol.~2, p.~5, 2015.

\bibitem{paper_4}
S.~T. Aguilar and V.~Jolivet, ``Handwritten text recognition for documentary
  medieval manuscripts,'' {\em Journal of Data Mining \& Digital Humanities},
  2023.

\bibitem{CILIA202339}
N.~D. Cilia, T.~D’Alessandro, C.~{De Stefano}, F.~Fontanella, and A.~{Scotto
  di Freca}, ``Comparing filter and wrapper approaches for feature selection in
  handwritten character recognition,'' {\em Pattern Recognition Letters},
  vol.~168, pp.~39--46, 2023.

\bibitem{paper_5}
N.~Rahal, L.~V\"{o}gtlin, and R.~Ingold, ``Historical document image analysis
  using controlled data for pre-training,'' {\em Int. J. Doc. Anal. Recognit.},
  vol.~26, p.~241–254, may 2023.

\bibitem{paper_6}
S.~Grieggs, C.~E.~M. Henderson, S.~Sobecki, A.~Gillespie, and W.~Scheirer,
  ``The paleographer's eye ex machina: Using computer vision to assist
  humanists in scribal hand identification,'' in {\em Proceedings of the
  IEEE/CVF Winter Conference on Applications of Computer Vision (WACV)},
  pp.~7177--7186, January 2024.

\bibitem{CILIA2020}
N.~Cilia, C.~{De Stefano}, F.~Fontanella, C.~Marrocco, M.~Molinara, and
  A.~{Scotto di Freca}, ``An end-to-end deep learning system for medieval
  writer identification,'' {\em Pattern Recognition Letters}, vol.~129,
  pp.~137--143, 2020.

\bibitem{paper_1}
L.~Lastilla, S.~Ammirati, D.~Firmani, N.~Komodakis, P.~Merialdo, and
  S.~Scardapane, ``Self-supervised learning for medieval handwriting
  identification: {A} case study from the vatican apostolic library,'' {\em
  Inf. Process. Manag.}, vol.~59, no.~3, p.~102875, 2022.

\bibitem{paper_3}
M.~Peer, F.~Kleber, and R.~Sablatnig, ``Towards writer retrieval for historical
  datasets,'' 2023.

\bibitem{Bulacu2007}
M.~Bulacu and L.~Schomaker, ``Text-independent writer identification and
  verification using textural and allographic features,'' {\em IEEE
  Transactions on Pattern Analysis and Machine Intelligence}, vol.~29, no.~4,
  pp.~701--717, 2007.

\bibitem{Marinai2020}
F.~Lombardi and S.~Marinai, ``Deep learning for historical document analysis
  and recognition—a survey,'' {\em Journal of Imaging}, vol.~6, no.~10, 2020.

\bibitem{DeSte18EAAI}
C.~{De Stefano}, M.~Maniaci, F.~Fontanella, and A.~{Scotto di Freca},
  ``Reliable writer identification in medieval manuscripts through page layout
  features: The avila bible case,'' {\em Engineering Applications of Artificial
  Intelligence}, vol.~72, pp.~99 -- 110, 2018.

\bibitem{CILIA2020_1}
N.~D. Cilia, C.~{De Stefano}, F.~Fontanella, M.~Molinara, and A.~{Scotto di
  Freca}, ``What is the minimum training data size to reliably identify writers
  in medieval manuscripts?,'' {\em Pattern Recognition Letters}, vol.~129,
  pp.~198--204, 2020.

\bibitem{DeSte18MEAS}
C.~{De Stefano}, M.~Maniaci, F.~Fontanella, and A.~{Scotto di Freca}, ``Layout
  measures for writer identification in mediaeval documents,'' {\em
  Measurement}, vol.~127, pp.~443 -- 452, 2018.

\bibitem{JIMG2020}
N.~D. Cilia, C.~De~Stefano, F.~Fontanella, C.~Marrocco, M.~Molinara, and
  A.~{Scotto di Freca}, ``An experimental comparison between deep learning and
  classical machine learning approaches for writer identification in medieval
  documents,'' {\em Journal of Imaging}, vol.~6, no.~9, 2020.

\bibitem{MT_icpr_25}
T.~D'Alessandro, C.~De~Stefano, F.~Fontanella, and A.~Scotto~di Freca, ``Writer
  identification in multiple medieval books: A preliminary study,'' in {\em
  Pattern Recognition} (A.~Antonacopoulos, S.~Chaudhuri, R.~Chellappa, C.-L.
  Liu, S.~Bhattacharya, and U.~Pal, eds.), (Cham), pp.~77--92, Springer Nature
  Switzerland, 2025.

\bibitem{CCN_tempMatc}
J.~Sarvaiya, S.~Patnaik, and S.~Bombaywala, ``Image registration by template
  matching using normalized cross-correlation,'' in {\em 2009 International
  Conference on Advances in Computing, Control, and Telecommunication
  Technologies}, pp.~819--822, 2009.

\bibitem{JIANG20221066}
P.~Jiang, D.~Ergu, F.~Liu, Y.~Cai, and B.~Ma, ``A review of yolo algorithm
  developments,'' {\em Procedia Computer Science}, vol.~199, pp.~1066--1073,
  2022.
\newblock The 8th International Conference on Information Technology and
  Quantitative Management (ITQM 2020 \& 2021): Developing Global Digital Economy
  after COVID-19.

\bibitem{YoloCH}
E.~S.~d. Santos~Júnior, T.~Paixão, and A.~B. Alvarez, ``Comparative
  performance of yolov8, yolov9, yolov10, and yolov11 for layout analysis of
  historical documents images,'' {\em Applied Sciences}, vol.~15, no.~6, 2025.

\bibitem{7}
M.~Maniaci and G.~Ornato, ``Prime considerazioni sulla genesi e la storia della
  bibbia di avila,'' in {\em Miscellanea F. Magistrale}, 2010.

\bibitem{AvilaB}
``{ BIBLIOTECA DIGITAL HISPÁNICA, Biblioteca Nacionalde España}.''

\bibitem{YOLO_1}
J.~Redmon, S.~K. Divvala, R.~B. Girshick, and A.~Farhadi, ``You only look once:
  Unified, real-time object detection,'' in {\em 2016 {IEEE} Conference on
  Computer Vision and Pattern Recognition, {CVPR} 2016, Las Vegas, NV, USA,
  June 27-30, 2016}, pp.~779--788, {IEEE} Computer Society, 2016.

\bibitem{Liu2022-wh}
H.~Liu, F.~Sun, J.~Gu, and L.~Deng, ``{SF-YOLOv5}: A lightweight small object
  detection algorithm based on improved feature fusion mode,'' {\em Sensors
  (Basel)}, vol.~22, p.~5817, Aug. 2022.

\bibitem{SPP_15}
K.~He, X.~Zhang, S.~Ren, and J.~Sun, ``Spatial pyramid pooling in deep
  convolutional networks for visual recognition,'' {\em IEEE Transactions on
  Pattern Analysis and Machine Intelligence}, vol.~37, no.~9, pp.~1904--1916,
  2015.

\bibitem{YOLOv5s6}
C.-L. Li and C.-Y. Su, ``Multi-connection of double residual block for yolov5
  object detection,'' in {\em 2022 8th International Conference on Applied
  System Innovation (ICASI)}, pp.~13--16, 2022.

\bibitem{coco}
T.-Y. Lin, M.~Maire, S.~Belongie, L.~Bourdev, R.~Girshick, J.~Hays, P.~Perona,
  D.~Ramanan, C.~L. Zitnick, and P.~Doll{\'a}r, ``Microsoft {COCO}: Common
  objects in context,'' 2014.

\bibitem{Otzu}
N.~Otsu, ``A threshold selection method from gray-level histograms,'' {\em IEEE
  Transactions on Systems, Man, and Cybernetics}, vol.~9, no.~1, pp.~62--66,
  1979.

\end{thebibliography}
\end{document}